\documentclass[a4paper, 12pt]{article}
\usepackage{graphicx}
\usepackage{amsfonts}

\makeatletter

\renewcommand{\ps@plain}{%
\renewcommand{\@oddfoot}{\footnotesize\rm The Electronic \footnotesize\sc Journal of
Symbolic Data Analysis
  {\footnotesize\rm - Vol. 2, N. 1} (2004), \hfil\rm\thepage}}
\makeatother
\title{self-organizing maps and symbolic data}

\author{A\"{\i}cha El Golli, Brieuc Conan-Guez, Fabrice Rossi\\
\small AxIS project, National Research Institute in Computer Science and Control (INRIA) \\[-0.8ex]
\small Rocquencourt Research Unit\\ [-0.8ex]
\small Domaine de Voluceau, Rocquencourt, \\[-0.8ex]
\small B.P. 105, 78153 Le Chesnay Cedex, France\\
\small \texttt{{aicha.elgolli, brieuc.conan-guez,
fabrice.rossi}@inria.fr}}

\date{\small Submitted: February 2004;  Accepted: September 2004}

\begin{document}
\maketitle

\begin{abstract}
In data analysis new forms of complex data have to be considered
like for example (symbolic data, functional data, web data, trees,
SQL query and multimedia data, ...). In this context classical
data analysis for knowledge discovery based on calculating the
center of gravity can not be used because input are not
$\mathbb{R}^p$ vectors.
In this paper, we present an application on real world symbolic data using the self-organizing map. To this end, 
we propose an extension of the self-organizing map that can handle
symbolic data.\\
\textbf{keywords:} Classification, Self organizing map, symbolic
data, dissimilarity.
\end{abstract}

\section{Introduction}
The self-organizing map(SOM) introduced by Kohonen \cite{Koh} is an unsupervised neural network method which has both clustering and visualization 
properties. It can be considered as an algorithm that maps a high dimensional data space, $\mathbb R^p$, to a lower dimension, generally 2, and which 
is called a map. This projection enables the input data to be partitioned into "similar" clusters while preserving their topology. Its most similar 
predecessors are the k-means algorithm \cite{Mac} and the dynamic clustering method \cite{Did}, which operate as a SOM without topology preservation 
and therefore without easy visualization.
In data analysis, new forms of complex data have to be considered, most notably symbolic data (data with an internal structure such as interval 
data, distributions, functional data, etc.) and semi-structured data (trees, XML documents, SQL queries, etc.). In this context, classical data 
analysis based on calculating the center of gravity can not be used because input are not $\mathbb R^p$ vectors. In order to solve this problem, 
several methods can be considered depending on the type of data (for example projection operators 
for functional data \cite{Ram}). However, those methods are not fully general and an adaptation of every data analysis algorithm to the resulting 
data is needed.

The Kohonen's SOM is based on the center of gravity notion and
unfortunately, this concept is not applicable to many kinds of
complex data. In this paper we propose an adaptation of the SOM to
dissimilarity data as an alternative solution. Our goal is to
modify the SOM algorithm to allow its implementation on
dissimilarity measures rather than on raw data. To this end, we
take one's inspiration from the work of Kohonen \cite{Koh96}. To
apply the method, only the definition of a dissimilarity for each
type of data is necessary and so complex data can be processed.

\section{Batch self-organizing map for dissimilarity data}
The SOM can be considered as carrying out vector quantization and/or clustering while preserving the spatial ordering of the prototype vectors (also 
called referent vectors) in one or two dimensional output space. The SOM consists of neurons organized on a regular low-dimensional map. More 
formally, the map is described by a graph $(C,\Gamma)$. $C$ is a set of $m$ interconnected neurons having a discrete topology defined by $\Gamma$.

For each pair of neurons (c, r) on the map, the distance $\delta(c, r)$,  is defined as the shortest path between c and r on the graph. This distance 
imposes a neighborhood relation between neurons. The batch training algorithm is an iterative algorithm in which the whole data set (noted $\Omega$) 
is presented to the map before any adjustments are made. We note $z_i$ an element of $\Omega$ and $\mathbf{z_i}$ the representation of this element 
in the space D called representation space of $\Omega$.  In our case, the main difference with the classical batch algorithm is that the 
representation space is not $\mathbb R^p$ but an arbitrary set on which dissimilarity (denoted d) is defined.

Each neuron c is represented by a set $A_c= {z_1,..., z_q}$ of
elements of $\Omega$ with a fixed cardinality $q$, where $z_i$
belongs to $\Omega$. $A_c$ is called an individual referent. We denote \emph{A} the set of all individual referents, i.e. the list \emph{$A = {A_1 ,...,A_m}$}. In our approach 
each neuron has a finite number of representations.
We define a new adequacy function \emph{$d^T$} from $\Omega \times P (\Omega)$ to $\mathbb R^+$ by:
$$ d^T(z_i, A_c)= \displaystyle\sum_{r\in C}K^T(\delta_{rc}) \sum_{z_j\in A_r}
d^2(\mathbf{z_i}, \mathbf{z_j} )$$ $d^T$ is based on the kernel
positive function $K$.  $K^T(\delta(c,r))$ is the neighborhood
kernel around the neuron r. This function is such that
$\displaystyle\lim_{\mid\delta\mid \longrightarrow
\infty}K(\delta)=0$
and allows us to transform the sharp graph distance between two neurons on the map $(\delta(c,r))$ into a smooth distance. $K$ is used to define a family of 
functions $K^T$ parameterized by T, with $k^T(\delta)=K(\frac{\delta}{T})$. T is used to control the size of the neighborhood \cite{Thi}: when the 
parameter T is small, there are few neurons in the neighborhood. A simple example of $K^T$ is defined by $K^T(\delta)=e^{-\frac{\delta^2}{T^2}}$.

During the learning, we minimize a cost function $E$ by
alternating an assignment step and a representation step. During
the assignment step, the assignment function $f$ assigns each
individual $z_i$ to the nearest neuron, here in terms of the
function $d^T$:
$$f(z_i)= arg \displaystyle\min_{c \in C} d^T(z_i,A_c)$$

If there is equality, we assign the individual $z_i$ to the neuron
with the smallest label.

During the representation step, we have to find the new individual
referents $A^*$ that represent the set of observations in the best
way in terms of the following cost function $E$:
$$E(f, A)= \sum_{z_i \in \Omega} d^T(z_i, A_{f(z_i)})
=\sum_{z_i \in \Omega} \sum_{r\in C}K^T(\delta(f(z_i), r))
\sum_{z_j\in A_r} d^2(\mathbf{z_i}, \mathbf{z_j} )$$ This function
calculates the adequacy between the induced partition by the
assignment function and the map referents $A$.

The criterion $E$ is additive so this optimization step can be carried out independently for each neuron. Indeed, we minimize the $m$ following 
functions:
$$E_r = \sum_{z_i \in \Omega} K^T(\delta(f(z_i), r)) \sum_{z_j\in A_r} d^2(\mathbf{z_i}, \mathbf{z_j} )$$

In the classical batch version, this minimization of $E$ function is immediate because the positions of the referent vectors are the averages of the 
data samples weighted by the kernel function.

\section{Experiments} To evaluate our method, we consider real world interval data. Our adaptation of the SOM to dissimilarity data is directly 
applied to this kind of interval structured data, once  we can associate dissimilarity to these data. This application concerns monthly minimal and 
maximal temperatures observed in 265 meteorological stations in China.  A natural representation of the monthly temperature recorded by a station is 
the interval constituted by the mean of the daily minimal and the mean of the daily maximal temperatures observed at this station over a month. 
Table~\ref{ChineSymbo} depicts the temperature recorded by the 265 stations over a 10-year period (between 1979 and 1988). Each interval is the mean 
of the minimal and the mean of the maximal monthly temperatures for these 10 years.

\begin{table}[htbp]
\centering
\begin{tabular}{|c|c|c|c|c|c|}\hline
Station& January&February& ...& November &December \\ \hline
Abag Qi &[-24.9; -17] & [-22.3; -12.8] & ... &[-16.4; -6.2] &[-24.7; -14.8] \\
$\vdots$& $\vdots$& $\vdots$& $\vdots$& $\vdots$& $\vdots$\\
Hailaer &[-28.6;  -22.5] &[-25.5;  -19.7] &...&[-17.4; -9.3] &[-25.5; -20.0]\\
$\vdots$& $\vdots$& $\vdots$& $\vdots$& $\vdots$& $\vdots$\\
\hline
\end{tabular}
\caption{Temperatures of the 265 Chinese stations between 1979 and 1988}
\label{ChineSymbo}
\end{table}
We will now describe the parameters used for this application (dissimilarity, map dimensions, number of iterations, ...). The choice of these 
parameters is important for the algorithm. We will then describe the obtained results. We use the factorial dissimilarity analysis (for more details \cite{Sap90}, \cite{elg04}) to 
visualize the maps.

\subsection{Hausdorff distance}
First, we choose to work with the Hausdorff-type L2-distance on
interval data defined as follows:
$$d(Q, Q')=\sqrt{\sum_{j=1}^p(max\{|a_j-a_j'|,|b_j,b_j'|\})^2}$$
with $Q=(I_1,...,I_p)$ and $Q'=(I_1',...,I_p')$ a pair of items described by $p$ intervals and $I_j=[a_j,b_j]$. It combines the $p$ one-dimensional, 
coordinate-wise Hausdorff distances in a way which is similar to the definition of the Euclidean distance in $\mathbb {R}^p$. The map dimension is 
$m=30$ neurons ($10 \times 3$). We use the elements of $\Omega$ in a random order to initialize the map and to choose the initial individual referents $A^0$. The 
cardinality of the individual referent $q$ is fixed to 1.

\begin{figure}[htbp]
\begin{center}
{\includegraphics[height=3cm]{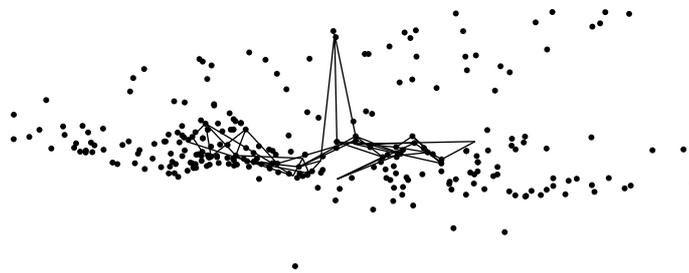}} \caption{Initial
map and the data on factorial dissimilarity analysis plan}
\label{Chineini}
\end{center}
\end{figure}

Figure \ref{Chineini} shows the initial map on factorial
dissimilarity analysis plans.\\

\begin{figure}[htbp]
\begin{center}
{\includegraphics[height=3cm]{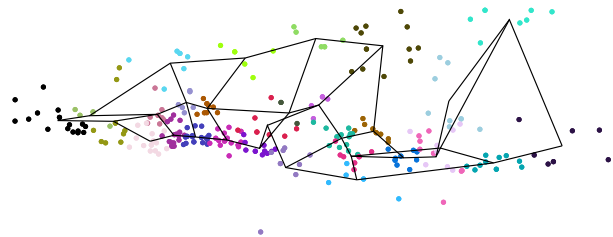}}
\caption{Final map and the data on factorial dissimilarity analysis plans. Each color represents a cluster}
\label{Chineout}
\end{center}
\end{figure}

Figure \ref{Chineout} shows the projection of the map that was 
finally obtained on the training data in factorial plans.\\

\begin{figure}[htbp]
\begin{center}
{\includegraphics[height=12cm, angle=-90]{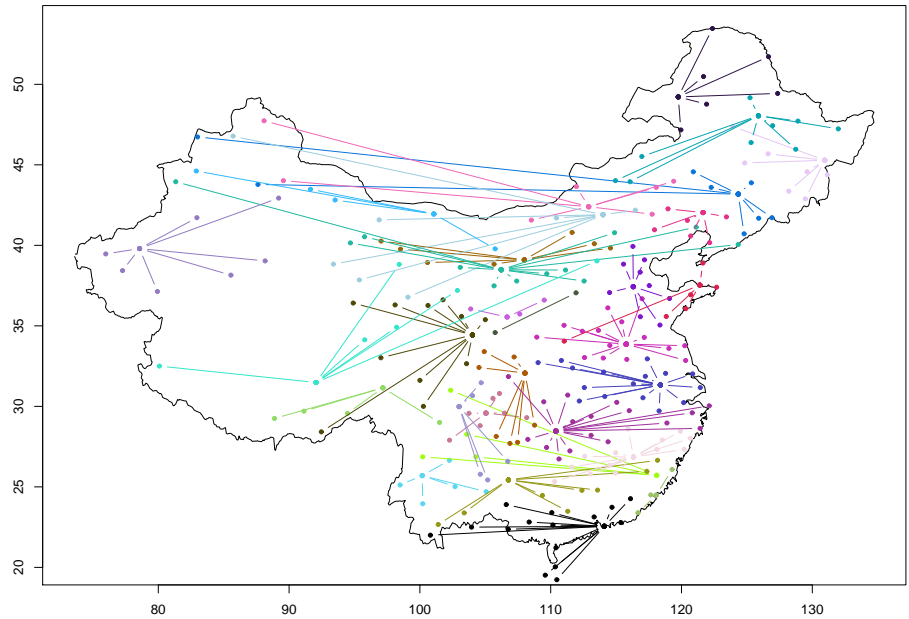}}
\caption{Distribution of the clusters on the geographical map of
China using the same colors of Figure 2} \label{Chine}
\end{center}
\end{figure}
The details of the result, shown in Figure \ref{Chine}, provide a nice representation of all 
the stations displayed over 30 clusters. These resulting clusters on the geographical map of China provide the representation of the stations 
attached to their referent station. The clusters on the right of Figure \ref{Chineout} are cold stations and correspond to the north and west of China. 
The warm stations are on the left of Figure \ref{Chineout} and correspond to the south and south-east of China and are characterized by very large 
variations in temperature. There is a continuity from cold stations to warm and hot ones. The analysis of the distribution of the clusters on the 
geographical map of China made it possible to deduce that the variations in temperature depend on latitude than on longitude.\\
\subsection{Euclidean distance}
Secondly, we use the Euclidean distance on interval data defined
as follows:

$$d(Q,Q')= 1/4 \|(a-a')+(b-b')\| ^2$$
with $Q=(I_1,...,I_p)$ and $Q'=(I_1',...,I_p')$ a pair of items
described by $p$ intervals and $I_j=[a_j,b_j]$.

We use the same parameters than for the Hausdorff distance.\\
\begin{figure}[htbp]
\begin{center}
\includegraphics[height=3cm]{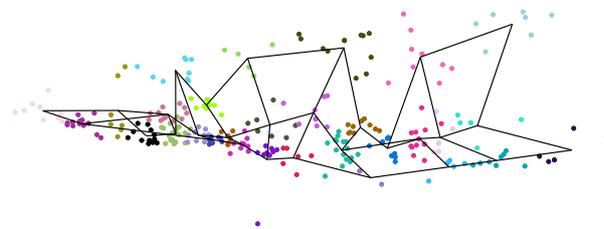}
\caption{Final map on factorial dissimilarity analysis plan. Each
color represent a cluster} \label{Echineout}
\end{center}
\end{figure}

\begin{figure}[htbp]
\begin{center}
{\includegraphics[height=12cm, angle=-90]{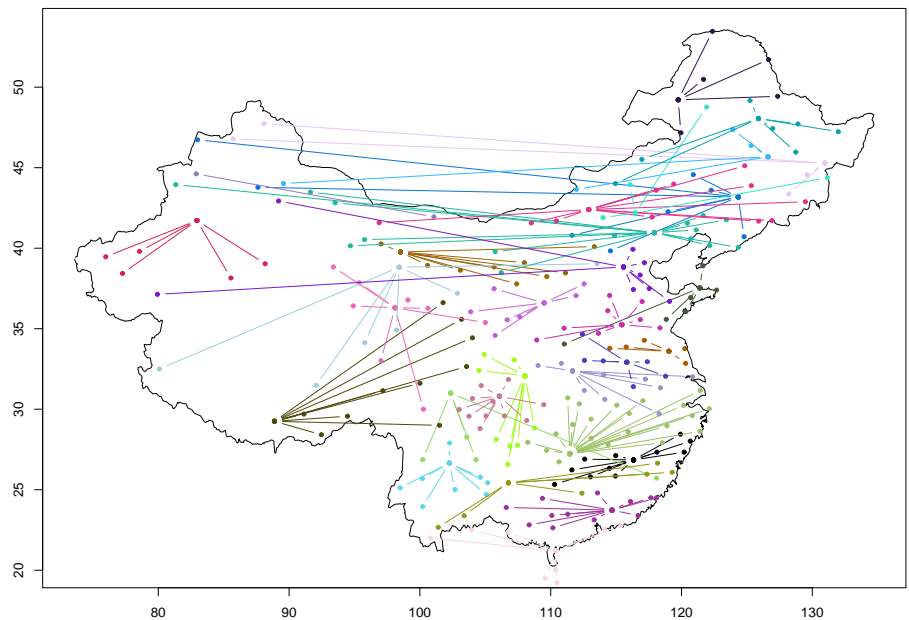}}
\caption{Distribution of the clusters on the geographical map of
China using the same colors of Figure 4} \label{Echine}
\end{center}
\end{figure}

Figure \ref{Echineout} shows the projection of the final map on
factorial dissimilarity analysis plan. Figure \ref{Echine}
provides the details of the result.

The classification is meaningful but different from the one with
the Hausdorff distance. We will now compare the different results.
\subsection{Discussion}
We use some other metrics for this application but we can't detail
the results for lack of space. We use the vertex-type distance
defined as the sum of the squared Euclidean distances between the
$2^p$ vertices. We use also the mean temperatures of the stations
(it's a non symbolic representation of the temperatures). In order
to compare the different results obtained by these different
metrics, we calculate longitude and latitude distortions of the
different obtained clustering. The distortion is defined as the
quadratic mean error between the referent and their assigned
individuals.

The longitudinal distortion is defined as follows:

$$(D_{long})^2= \sum_{c \in C}\sum_{z_i \in c}\frac{1}{|c|} |Lo_{z_i}-Lo_{f(z_i)}|^2$$ with $|c|$ the cardinal of the cluster $c$, $|Lo_{z_i}- 
Lo_{f_(z_i)}|$ the longitude distance between the station $z_i$
and his referent $f(z_i)$

The latitude distortion is defined as follows:

$$(D_{lati})^2= \sum_{c \in C}\sum_{z_i \in c}\frac{1}{|c|} |La_{z_i}-La_{f(z_i)}|^2$$ with $|La_{z_i}-La_{f(z_i)}|^2$ the latitude distance 
between the station $z_i$ and his referent $f(z_i)$.

In the table \ref{comp1}, we represent the longitude and the latitude distortions of the different obtained clustering with the 
different metrics.

\begin{table}[htbp]
\begin{center}
\begin{tabular}{|c|c|c|c|}\hline

 Data type &Used metric& Longitude distortion & Latitude distortion \\ \hline
\hline Intervals& Euclidean distance& 9.250688&  1.993213  \\
\hline Intervals& Vertex-type distance  &8.625175&  2.165838 \\
\hline Means(numerics) &Euclidean distance& 7.656033& 1.936692 \\
\hline Intervals&Hausdorff distance& 7.38314& 1.911461\\ \hline

\end{tabular}
\end{center}
\caption{The different longitude and latitude distortions for the
different metrics} \label{comp1}
\end{table}

We can deduce that the clustering obtained with the Hausdorff
distance induced the smallest latitude and longitude distortions.

\section{Conclusion} In this paper, we proposed an adaptation of the self-organizing map to dissimilarity data. This adaptation is based on the batch 
algorithm and can handle both numerical data and complex data. The experiments showed the usefulness of the method and that it can be applied to 
symbolic data or other complex data once we can define
dissimilarity for these data.

\end{document}